\documentclass{llncs}
\usepackage{amsmath}
\usepackage{bm}
\usepackage[ruled,linesnumbered]{algorithm2e}
\usepackage{booktabs}
\usepackage{multirow}
\usepackage[misc]{ifsym}
\usepackage{graphicx}
\usepackage{subfig}
\usepackage{hyperref}
\usepackage{color}

\begin{document}
\title{Floorplanning of VLSI by Mixed-Variable Optimization}
\author{Jian Sun\inst{1}\and Huabin Cheng\inst{2} \and Jian Wu\inst{3} \and Zhanyang Zhu\inst{3} \and Yu Chen\inst{1}$^{(\textrm{\Letter})}$}
\institute{School of Science, Wuhan University of Technology, Wuhan, 430070, China \email{ychen@whut.edu.cn}\and Department of Basic Science, Wuchang Shouyi University, Wuhan, 430064, China \and School of Information Engineering, Wuhan University of Technology, Wuhan, 430070, China}

\maketitle
\begin{abstract}
By formulating the floorplanning of VLSI as a mixed-variable optimization problem, this paper proposes to solve it by memetic algorithms, where the discrete orientation variables are addressed by the distribution evolutionary algorithm based on a population of probability model (DEA-PPM), and the continuous coordination variables are optimized by the conjugate sub-gradient algorithm (CSA). Accordingly, the fixed-outline floorplanning algorithm based on CSA and DEA-PPM (FFA-CD) and the floorplanning algorithm with golden section strategy (FA-GSS) are proposed for the floorplanning problems with and without fixed-outline constraint. 
Numerical experiments on GSRC test circuits show that the proposed algorithms are superior to some celebrated B*-tree based floorplanning algorithms, and are expected to be applied to large-scale floorplanning problems due to their low time complexity.\\
	
\end{abstract}
\section{Introduction}
Floorplanning is a critical stage in the physical design of very large-scale integration circuit (VLSI) that determines the performance of VLSI chips to a large extent~\cite{clar:eke1}. It is a complex optimization problem with multiple objectives and constraints, which makes it challenging to develop high-performance algorithms for floorplanning of VLSI~\cite{clar:eke2}. 

Floorplanning algorithms generally fall into two categories: the floorplanning algorithm based on combinatorial optimization model (FA-COM) and the floorplanning algorithm based on analytic optimization model (FA-AOM). Representing the relative positions of macros by combinatorial coding structures such as the B*-tree, the sequential pair, etc., one can formulate the floorplanning problem as a combinatorial optimization problem, which is then addressed by metaheuristics in the FA-COMs~\cite{clar:eke3,clar:eke4,clar:eke5,clar:eke6}. 
The combinatorial codes representing relative positions of macros can be naturally decoded into the compact floorplans complying with the non-overlapping constraints, however, the combinatorial explosion contributes to poor performances of FA-COM on large-scale cases. Accordingly, the problem size could be reduced by clustering or partitioning strategies, which in turn makes it hard to converge to the global optimal results of the investigated large-scale floorplanning problems~\cite{clar:eke7,clar:eke8}.

FA-AOMs address analytical floorplanning models by continuous optimization algorithms, which contributes to their lower time complexities on large-scale cases~\cite{clar:eke9,clar:eke10}.
Since the optimization results of continuous optimization algorithms do not fulfill the non-overlapping constraints for most cases, a FA-AOM usually consists of the global floorplanning stage and the legalization stage, the first optimizing the overall evaluation index, and the second tuning the positions of macros to eliminate constraint violations of results. Li \emph{et al.}~\cite{clar:eke11} proposed an analytic floorplanning algorithm for large-scale floorplanning cases, where the fixed-outline global floorplanning was implemented by optimizing the electrostatic field model of global placement.  In the legalization stage, horizontal constraint graphs and vertical constraint graphs were constructed to eliminate overlap of floorplanning results. Huang \emph{et al.}~\cite{Huang2023} presented an improved electrostatics-based analytical method for fixed-outline floorplanning, which incorporates module rotation and sizing driven by wirelength.

Since some of the evaluation indexes of global floorplanning are not smooth, additional smooth approximation to the optimization objective function could be incorporated to achieve fast convergence of gradient-based optimization algorithms. However, the approximation procedure not only introduces extra time complexity of the FA-AOM, but also leads to its local convergence to an optimal solution significantly different from that of the original non-smooth model. Accordingly, the conjugate subgradient algorithm~\cite{clar:eke12} is employed in this paper to deal with the continuous variables representing coordinates of modules. Meanwhile, we address the orientation of modules  by discrete variables, and formulate the floorplanning problem as a mixed-variable optimization problem.

Rest of this paper is organized as follows. Section \ref{sec:Pre} introduces some preliminaries. Then, the proposed  algorithms developed for floorplanning problems with and without fixed-outline constraints are presented in Sections \ref{sec:FFA-CD} and \ref{sec:FA-GSS}, respectively. Numerical experiment is performed in Section \ref{sec:Num} to demonstrate the competitiveness of the proposed algorithms, and Section \ref{sec:Con} concludes this paper.
%
%
%
\section{Preliminaries}\label{sec:Pre}
\subsection{Problem Statement}
Given a collection of rectangular modules $V = \{ v_1, v_2, \ldots, v_n \}$ and a set of edges (networks) $E = \{ e_1, e_2, \ldots, e_m \}$, the VLSI floorplanning problem tries to minimize the total wirelength and the floorplan area by placing modules in approximate positions. Denote the center coordinates of module $v_i$ be $(x_i,y_i)$, and its orientation is represented by $r_i$.  A floorplan of VLSI is represented by the combination of vectors $\bm{x}$, $\bm{y}$ and $\bm{r}$, where $\bm{x}=(x_1,x_2,\dots,x_n)$, $\bm{y}=(y_1,y_2,\dots,y_n)$, $\bm{r}=(r_1,r_2,\dots,r_n)$.
Subject to the constraint of placing non-overlaping modules with a fixed outline, the floorplanning problem is formulated as
\begin{equation}\label{pro:cons}
	\begin{aligned}
		&\min \quad  W(\bm x,\bm y) \\
		& s.t.\quad \left\{\begin{array}{l}
			 D(\bm x,\bm y, \bm r)=0, \\
			B(\bm x,\bm y, \bm r)=0,
		\end{array} \right.
	\end{aligned}
	\end{equation}
where $W(\bm x,\bm y)$ is the total wirelength, 
$D(\bm x,\bm y, \bm r)$ is the sum of overlapping area, and $B(\bm x,\bm y, \bm r)$ is the sum of width beyond the fixed outline. By the Lagrange multiplier method, it can be transformed into an unconstrained optimization model
\begin{equation}\label{pro:gf}
		\min \,\, f(\bm x,\bm y, \bm r) = \alpha W(\bm x,\bm y) + \lambda \sqrt{D(\bm x,\bm y, \bm r)} + \mu B(\bm x,\bm y, \bm r),
	\end{equation}
where $\alpha$, $\lambda$, and $\mu$ are parameters to be confirmed. Here, the square root of  $D(\bm x,\bm y, \bm r)$ is adopted to ensure that all indexes to be minimized are of the same dimension.
	
\paragraph{Total Wirelength $W(\bm x,\bm y)$}: The total wirelength is here taken as the total sum of half-perimeter wirelength (HWPL)
	\begin{equation}\label{obj:HWPL}
		W(\bm x,\bm y)=\sum_{e\in E} (\max\limits_{v_i \in e}  x_i-\min\limits_{v_i \in e}  x_i +\max\limits_{v_i \in e}  y_i-\min\limits_{v_i \in e}  y_i)
	\end{equation}
	
	
\paragraph{Sum of Overlapping Area $D(\bm x,\bm y, \bm r)$}: The sum of overlapping area is computed by
\begin{equation}\label{obj:SOA}
		D(\bm x,\bm y,\bm r)=\sum_{i,j}O_{i,j}(\bm x,\bm r) \times O_{i,j}(\bm y, \bm r),
\end{equation}
where $O_{i,j}(\bm x,\bm r)$ and $O_{i,j}(\bm y,\bm r)$ represent the overlapping lengths of modules $i$ and $j$ in the $X$-axis and $Y$-axis directions, respectively. Denoting $\Delta_x(i,j)=|x_i-x_j|$, we know
\begin{equation}\label{ind:OLX}
O_{i,j}(\bm x, \bm r)=
	\begin{cases}
		\max(\hat w_i,\hat w_j), & \mbox{if } 0\le \Delta_x(i,j) \le \frac{|\hat w_i-\hat w_j|}{2},\\
		\frac{\hat w_i-2\Delta_x(i,j)+\hat w_j}{2}, & \mbox{if } \frac{|\hat w_i-\hat w_j|}{2} < \Delta_x(i,j) \le \frac{\hat w_i+\hat w_j}{2},\\
		0, &  \mbox{if }  \frac{\hat w_i+\hat w_j}{2} < \Delta_x(i,j) ,
	\end{cases}
\end{equation}
where $\hat{w}_i$ is confirmed by
\begin{equation}\label{para:width}
  \hat{w}_i=\begin{cases}
              w_i, & \mbox{if } r_i\in\{0,\pi\}, \\
              h_i, & \mbox{otherwise},
            \end{cases}\quad i\in\{1,2\dots,n\}.
\end{equation}
 Denoting $\Delta_y(i,j)=|y_i-y_j|$, we have
\begin{equation}\label{ind:OLY}
O_{i,j}(\bm y, \bm r)=
	\begin{cases}
		\max(\hat h_i,\hat h_j), & \mbox{if } 0\le \Delta_y(i,j) \le \frac{|\hat h_i-\hat h_j|}{2},\\
		\frac{\hat h_i-2\Delta_y(i,j)+\hat h_j}{2}, & \mbox{if } \frac{|\hat h_i-\hat h_j|}{2} < \Delta_y(i,j) \le \frac{\hat h_i+\hat h_j}{2},\\
		0, &  \mbox{if }  \frac{\hat h_i+\hat h_j}{2} < \Delta_y(i,j) ,
	\end{cases}
\end{equation}
where $\hat{h}_i$ is confirmed by
\begin{equation}\label{para:height}
  \hat{h}_i=\begin{cases}
              h_i, & \mbox{if } r_i\in\{0,\pi\}, \\
              w_i, & \mbox{otherwise},
            \end{cases}\quad i\in\{1,2\dots,n\}.
\end{equation}

\paragraph{Sum of Width beyond the Fixed Outline $B(\bm x,\bm y, \bm r)$}: For floorplanning problems with fixed-outline, the positions of modules must meet the following constraints:
	\begin{equation*}
		\begin{cases}
			0 \le x_i-\hat w_i/2, \quad x_i+\hat w_i/2 \le W^*, \\
			0 \le y_i-\hat h_i/2, \quad y_i+\hat h_i/2 \le H^*,
		\end{cases}
	\end{equation*}
where $W^*$ and $H^*$ are the width and the height of square outline, respectively. Let
	\begin{align*}
		&b_{1,i}(\bm x)=\max (0,\hat w_i/2-x_i),b_{2,i}(\bm x)=\max (0,\hat w_i/2+x_i-W^*), \\		&b_{1,i}(\bm y)=\max (0,\hat h_i/2-y_i),b_{2,i}(\bm y)=\max 0,\hat h_i/2+y_i-H^*),
	\end{align*}
$\hat{w}_i$ and $\hat{h}_i$ are confirmed by (\ref{para:width}) and (\ref{para:height}), respectively.
Accordingly, $B(\bm x,\bm y,\bm r)$ can be confirmed by
\begin{equation}\label{obj:SWFO}
B(\bm x,\bm y,\bm r)=\sum_{i=1}^{n}(b_{1,i}(\bm x)+b_{2,i}(\bm x)+b_{1,i}(\bm y)+b_{2,i}(\bm y)),
\end{equation}
which is smoothed by
\begin{equation}\label{obj:SSWFO}
\widetilde{B}(\bm x,\bm y,\bm r)=\sum_{i=1}^{n}(b_{1,i}^2(\bm x)+b_{2,i}^2(\bm x)+b_{1,i}^2(\bm y)+b_{2,i}^2(\bm y)).
\end{equation}
	
	Let $\beta=0$, we get  for legitimization of the global floorplanning result the optimization problem
	\begin{equation}\label{pro:l}
		\min\quad \tilde{f}(\bm x,\bm y,\bm r)=\lambda_0 D(\bm x,\bm y,\bm r)+\mu_0 \widetilde{B}(\bm x,\bm y,\bm r).
	\end{equation}

\subsection{The Conjugate Sub-gradient Algorithm for Optimization of the Coordinate}
\begin{algorithm}[htb]
		\caption{$\bm{u}^*=CSA(f,\bm{u}_0,k_{max},s_0)$}\label{alg:CSA}
		\KwIn{Objective function $f(\bm{u})$, Initial solution $\bm{u}_0$, Maximum iterations $k_{max}$, Initial step control parameter $s_0$;}
		\KwOut{Optimal solution $\bm{u}^*$;}
		$\bm{g}_0 \in \partial f(\bm{u}_0)$, $\bm{d}_0=\bm{0}$, $k \leftarrow 1$\;
		\While{termination-condition 1 is not satisfied}
		{calculated subgradient $\bm{g}_k \in \partial f(\bm{u}_{k-1})$\;
			calculate Polak-Ribiere parameters $\eta _k=\frac{\bm{g}_k^T (\bm{g}_k-\bm{g}_{k-1})}{||\bm{g}_{k-1}||_2^2}$\;
			computed conjugate directions $\bm{d}_k=-\bm{g}_k+\eta _k \bm{d}_{k-1}$\;
			calculating step size $a_k=s_{k-1}/||\bm{d}_k||_2$\;
			renewal solution $\bm{u}_k=\bm{u}_{k-1}+a_k \bm{d}_k$\;
			update step control parameters $s_k=qs_{k-1}$\;
			updata $\bm{u}^*$\;}
	\end{algorithm}
	Zhu et al.\cite{clar:eke12} proposed to solving the non-smooth continuous optimization model of the global placement by the conjugate sub-gradient algorithm (CSA). With an initial solution $\bm{u}_0$, the pseudo code of CSA is presented in Algorithm \ref{alg:CSA}. Because the CSA is not necessarily gradient-descendant, the step size  has a significant influence on its convergence performance. The step size is determined by the norm of the conjugate directions together with the control  parameter $s_k$, which is updated as $s_k=qs_{k-1}$. As an initial study, we set $q=0.997$ in this paper. The \emph{termination-condition 1} is satisfied if $k$ is greater than a given budget $k_{max}$ or several consecutive iterations fails to get a better solution.


\subsection{The Distribution Evolutionary Algorithm for Optimization of the Orientation}
Besides the coordinate vectors $\bm{x}$ and $\bm{y}$, the floorplan is also confirmed by the orientation vectors $\bm{r}$. The orientation of modules is confirmed by clockwise rotation, and we set $r_i=j$ if the rotation angle is $\theta_i=j\pi/2$, $j=0,1,2,3$, $i=1,\dots,n$. The, optimization of the orientation vectors contributes to a combinatorial optimization problem.


The estimation of distribution algorithm (EDA) is a kind of metaheuristics that can address the combinatorial optimization problem well, but its balance between global exploration and local exploitation is a challenging issue~\cite{clar:eke13}. Xu \emph{et al.}~\cite{clar:eke14} proposed  for the graph coloring problem a distribution evolutionary algorithm based on a population of probability model (DEA-PPM), where a novel probability model and the associated orthogonal search are introduced to achieve well convergence performance on large-scale combinatorial problems.
	
The core idea of DEA-PPM for floorplanning is to simulate the probability distribution of orientations by constructing a probability matrix 
	\begin{equation}
		\bm{q}=(\vec{q}_1,\dots,\vec{q}_n)=
		\begin{bmatrix}
			q_{11}	&	q_{12} \cdots & q_{1n} \\
			q_{21}	&	q_{22} \cdots & q_{2n} \\
			q_{31}	&	q_{32} \cdots & q_{3n} \\
			q_{41}	&	q_{42} \cdots & q_{4n}
		\end{bmatrix},
	\end{equation}
where $\vec{q}_{j}$ representing the probability that module $j$ satisfies
	\begin{equation}
		||\vec{q}_j||_2^2=\sum_{i=1}^{k}q_{ij}^2=1,\quad \forall j=1,\dots,n.
	\end{equation}
Then, the random initialization of $\bm{q}$ generates a distribution matrix
	\begin{equation}\label{ini:q}
		\bm{q}{(0)}=
		\begin{bmatrix}
			{1}/{2}	&	{1}/{2}	&	\cdots	& {1}/{2} \\
			{1}/{2}	&	{1}/{2}	&	\cdots	& {1}/{2} \\
			{1}/{2}	&	{1}/{2}	&	\cdots	& {1}/{2} \\
			{1}/{2}	&	{1}/{2}	&	\cdots	& {1}/{2}
		\end{bmatrix}.
	\end{equation}
	
	The implementation of DEA-PPM is based on distributed population $\bm{Q}(t)=(\bm{q}^{[1]},\dots,\bm{q}^{[np]})$ and solution population $\bm{P}(t)=(\bm{p}^{[1]},\dots,\bm{p}^{[np]})$, which are employed here for the probability distributions and instantiations of orientation, respectively. Global convergence of DEA-PPM is achieved by an orthogonal search on $\bm{Q}(t)$, and the local exploitation  are implemented in both the distribution space and the solution space.

\section{The Fixed-outline Floorplanning Algorithm Based on CSA and DEA-PPM}\label{sec:FFA-CD}
\subsection{Framework}
	In this paper, the fixed-outline floorplanning algorithm based on CSA and DEA-PPM (FFA-CD) is proposed to solve the problem of fixed-outline floorplanning, where the DEA-PPM is employed to optimize the orientations of the modules and the CSA is used to optimize the corresponding coordinates of the modules.
	

The framework of FFA-CD is presented in Algorithm \ref{alg:FFA-CD}. It starts with initialization of the distribution and solution populations $\bm{Q}(0)$ and $\bm{P}(0)$, where $\bm{P}(0)$ consists of orientation combinations of modules. Meanwhile, the corresponding population $\bm{X}$ and $\bm{Y}$ of module coordinate is initialized by Latin hypercube sampling~\cite{clar:eke15}. Combining the orientation and coordinates of modules, we get the best coordinate vectors $\bm{x}^*$ and $\bm{y}^*$, as well as the corresponding orientation vector $\bm{r}^*$. Then, the \emph{while loop} of DEA-PPM is implemented to update $\bm{Q}(t)$ and $\bm{P}(t)$, where the CSA is deployed in \emph{UpdateXY} to get the best module coordinate.

\begin{algorithm}[htb]
		\caption{FFA-CD}\label{alg:FFA-CD}
		\KwIn{$f(\bm{x},\bm{y},\bm{r})$, $\widetilde{f}(\bm{x},\bm{y},\bm{r})$.}
		\KwOut{Optimal coordinate vector $(\bm{x}^*,\bm{y}^*)$ and  orientation vector $\bm{r}^*$.}

        {initialize the step control parameter $s$}\;
		initialize $\bm{Q}(0)$ by (\ref{ini:q}), and generate $\bm{P}(0)$ by sampling $\bm{Q}(0)$\;
		initialize $\bm{X}$ and $\bm{Y}$ by Latin hypercube sampling\;

		let $$(\bm{x}^*,\bm{y}^*,\bm{p}^*)=\arg\min f(\bm{x},\bm{y},\bm{r}),\bm{x}\in\bm{X},\bm{y}\in\bm{Y},\bm{r}\in\bm{P}(0);$$
set $\bm{q}^*$ as the distribution $\bm{q}$ corresponding to $\bm{p}^*$\;
		{set $t\leftarrow 1$, $\alpha \leftarrow 1$, $\lambda \leftarrow 20$, $\mu \leftarrow 100$, $\lambda_0 \leftarrow 1$, $\mu_0 \leftarrow 10$, $k_{\mbox{max}} \leftarrow 50$}\;
		\While{termination-condition 2 is not satisfied}
		{$\bm{Q}'(t)=OrthExpQ(\bm{Q}(t-1),\bm{P}(t-1))$;
			$\bm{P}'(t)=SampleP(\bm{Q}(t),\bm{P}(t-1))$;
			$(\bm{P}(t),\bm{X},\bm{Y},s)=UpdateXY(\bm{P}'(t),\bm{X},\bm{Y},s)$;
			$\bm{Q}(t)=RefineQ(\bm{P}'(t),\bm{P}(t),\bm{Q}'(t))$;
			$t=t+1$;}
	\end{algorithm}

\subsection{Evolution of the Distribution Population}
	In order to better explore the distribution space, DEA-PPM carries out orthogonal exploration for individuals in $\bm{Q}(t)$. 	
	Algorithm \ref{alg:OrthExp} gives the flow of orthogonal exploration, which aims to change $m$ worst individuals in $\bm{Q}$ by orthogonal transformation performed on $c$ columns of a distribution matrix. Here, $m$ is a random integer in $[1,np/2]$ and $c$ is a random integer in $[1,n/10]$.
	
\begin{algorithm}[htb]
		\caption{$\bm{Q}'=OrthExpQ(\bm{Q},\bm{P})$}\label{alg:OrthExp}
		\KwIn{$\bm{Q},\bm{P}$;}
		\KwOut{$\bm{Q}'$;}
		sorting $\bm{Q}$ by fitness values of corresponding individuals of $\bm{Q}$\;
		take $\bm{Q}_w$ as the collection of $m$ worst individuals of $\bm{Q}$\;
		$\bm{Q}'=\bm{Q} \setminus \bm{Q}_w$\;
		\For{$\bm{q} \in \bm{Q}_w$}
		{$\bm{q}' \leftarrow \bm{q}$\;
		randomly select $c$ columns $\vec{q}_{j_l}'(l=1,\dots,c)$ from $\bm{q}'$\;
			\For{l=1,\dots,c}
			{generate a random orthogonal matrix $\bm{M}_l$\;
			$\vec{q}_{jl}'=\bm{M}_l \vec{q}_{jl}'$;}
		$\bm{Q}'=\bm{Q}' \cup \bm{q}'$
		}
	\end{algorithm}

\begin{algorithm}[htb]
		\caption{$\bm{Q}'=RefineQ(\bm{P}',\bm{P},\bm{Q}')$}\label{alg:RefQ}
		\KwIn{$\bm{P}',\bm{P},\bm{Q}'$;}
		\KwOut{$\bm{Q}$;}
		\For{$i=1,\dots,np$}
		{$\bm{q}^{[i]} \in \bm{Q}',\bm{v}'^{[i]} \in \bm{P}',\bm{v}^{[i]} \in \bm{P}$\;
			\For{j=1,\dots,n}
			{set $rnd_j \sim U(0,1)$\;
				\eIf{$rnd_j \le p_0$}{$\vec{r}_j^{[i]}$ is generated by the exploitation strategy (Eqs. (\ref{str:exp1}) and (\ref{str:exp2}));}{$\vec{r}_j^{[i]}$ is generated by the disturbance strategy (Eq. (\ref{str:Dist}));}
			}
		$\bm{r}^{[i]}=(\vec{r}_1^{[i]},\dots,\vec{r}_n^{[i]})$;
		}
	\end{algorithm}
	
	In Algorithm \ref{alg:RefQ}, the intermediate distribution population $\bm{Q}'(t)$ is further updated to get $\bm{Q}(t)$. Given $\bm{q}^{[i]} \in \bm{Q}'(t)$,  it is updated using $\bm{v}'^{[i]}$ and $\bm{v}^{[i]}$, two orientation combinations selected from $\bm{P}'$ and $\bm{P}$, respectively. Columns of $\bm{q}^{[i]}$ are updated using either a exploitation strategy or a disturbance strategy presented as follows.
	
\paragraph{The exploitation strategy:} To update the $j^{th}$ column of $\bm{q}^{[i]}$, it is first renewed as
	\begin{equation}\label{str:exp1}
		r_{l,j}^{[i]}=\left\{
		\begin{array}{lcl}
			\sqrt{\alpha_0+(1-\alpha_0)(q^{[i]}_{l,j})^2}, &      &\mbox{if } l=v_j^{[i]}\\
			\sqrt{(1-\alpha_0)(q^{[i]}_{l,j})^2}, &      & \mbox{if } l \neq v_j^{[i]}
		\end{array} \right.
		\quad l=1,\dots,4,
	\end{equation}
	where $v_j^{[i]}$ is the $j^{th}$ component of $\bm{v}^{[i]}$. Then, an local orthogonal transformation is performed as
	\begin{equation}\label{str:exp2}
		\begin{bmatrix}
			r_{l_1,j}^{[i]}  \\
			r_{l_2,j}^{[i]}
		\end{bmatrix}	
		=U(\Delta \theta_j) \times
		\begin{bmatrix}
			r_{l_1,j}^{[i]}  \\
			r_{l_2,j}^{[i]}
		\end{bmatrix},	
	\end{equation}
	where $l_1=v_j'^{[i]}$, $l_2=v_j^{[i]}$. $U(\Delta \theta_j)$ is an orthogonal matrix given by
	\begin{equation*}
		U(\Delta \theta_j)=
		\begin{bmatrix}
			\cos(\Delta \theta_j)	&	  -\sin(\Delta \theta_j)\\
			\sin(\Delta \theta_j)	&	\cos(\Delta \theta_j)
		\end{bmatrix}.	
	\end{equation*}
	
\paragraph{The disturbance strategy:} In order to prevent the distribution population from premature, the disturbance strategy is performed as
	\begin{equation}\label{str:Dist}
		r_{l,j}^{[i]}=\left\{
		\begin{array}{rcl}
			\frac{\lambda (q^{[i]}_{l_0,j})^2}{1-(1-\lambda)(q^{[i]}_{l_0,j})^2}, &      &\mbox{if } l=l_0\\
		\frac{(q^{[i]}_{l,j})^2}{1-(1-\lambda)(q^{[i]}_{l_0,j})^2}, &      &\mbox{if } l \neq l_0
		\end{array} \right.
	\end{equation}
	where $l_0=v_j^{[i]}$.

\subsection{Optimization of the Floorplan with a Fixed Outline}
The floorplan is represented by the orientation vector $\bm{r}$ and the coordinate vectors $\bm{x}$ and $\bm{y}$. In FFA-CD, the evolution of orientation vectors is implemented by iteration of solution population $\bm{P}(t)$, and the corresponding coordinate vectors are optimized by the function \emph{UpdateXY}.	
\subsubsection{Initialization of the module orientation} According to the principle of DEA-PPM, the solution population $\bm{P}'(t)$ is obtained by sampling the distribution population $\bm{Q}'(t)$. To accelerate the convergence process, the sampling process is performed with inheritance as the process illustrated in Algorithm \ref{alg:Samp}.
	\begin{algorithm}[htb]
		\caption{$\bm{P}'=SampleP(\bm{Q},\bm{P})$}\label{alg:Samp}
		\KwIn{$\bm{Q},\bm{P}$;}
		\KwOut{$\bm{P}'$;}
		\For{$i=1,\dots,np$}
		{$\bm{q}^{[i]} \in \bm{Q},\bm{v}^{[i]} \in \bm{P}$\;
			\For{j=1,\dots,n}
			{set $rnd_j \sim U(0,1)$\;
				\eIf{$rnd_j \le r$}{sampling $\vec{q}_j^{[i]}$ to get $\vec{v}_j'^{[i]}$;}{$\vec{v}_j'^{[i]}=\vec{v}_j^{[i]}$;}
			}
			$\bm{v}'^{[i]}=(\vec{v}_1'^{[i]},\dots,\vec{v}_n'^{[i]})$;
		}
	$\bm{P}'=\bigcup_{i=1}^{np}\bm{v}'^{[i]}$
	\end{algorithm}

\subsubsection{Optimization of module position}
With the orientation of modules confirmed by the solution population, the position of the modules is optimized by Algorithm \ref{alg:UpdXY}. For a combination of position vector $(\bm{x}^{[i]},\bm{y}^{[i]})$, the global floorplanning is first implemented by optimizing $f$; then, the weights of the constraint items is increased to legalize the floorplan approach by lines 4-7, or the legalization process is implemented by lines 9-10. The legalization process based on constraint graphs~\cite{clar:eke10} are implemented \emph{Graph()}, which is presented in Algorithm \ref{alg:Graph}. To prevent $\bm{X}$ and $\bm{Y}$ from falling into inferior local solutions, the coordinates are reinitialized if no better solution is obtained for several times.

\begin{algorithm}[htb]		\caption{$(\bm{P}(t),\bm{X},\bm{Y},s)=UpdateXY(\bm{P}'(t),\bm{X},\bm{Y},s)$}\label{alg:UpdXY}
		\KwIn{$\bm{X},\bm{Y},s$;}
		\KwOut{$\bm{X},\bm{Y},s$;}
		\For{$i=1,\dots,np$}
		{
$(\bm{x}^{[i]},\bm{y}^{[i]})=CSA(f,(\bm{x}^{[i]},\bm{y}^{[i]}),k_{\mbox{max}},s)$\;
			\eIf{$d_0>\delta_1$}
			{
				$\lambda$=$\mathop{min}(1.5\lambda,\lambda+30)$\;
				\If{$c_0>\delta_2$}
				{
					$\mu$=$\mathop{min}(1.1\mu,\mu+10)$;
				}
			}{$(\bm{x}^{[i]},\bm{y}^{[i]})=CSA(\widetilde{f},(\bm{x}^{[i]},\bm{y}^{[i]}),1000,\mathop{max}(s/2,50))$\;
			$(\bm{x}^{[i]},\bm{y}^{[i]})=Graph(\bm{x}^{[i]},\bm{y}^{[i]})$;}
		}
		$s=\mathop{max}(0.95*s,s_{\mbox{min}})$\;
		\If{no better solution is obtained for several times
		}{reinitialize $\bm{X},\bm{Y}$;}
\end{algorithm}
	
	The legalization of Algorithm \ref{alg:Graph} is implemented as follows.  Let $(x_i',y_i')$ be the lower-left coordinate of block $v_i$. $v_i$ is \emph{to the left of} $v_j$ if it holds
$$O_{i,j}(y) > 0, O_{i,j}(x)=0, x_i'<x_j';$$
 $v_i$ is \emph{to the below of} $v_j$ if
$$O_{i,j}(x) > 0, O_{i,j}(y)=0, y_i'<y_j'.$$
Denote $I_i$ and $J_i$ as the left-module set and the lower-module set of module $i$, respectively. Then, the $x$- and $y$-coordinates of module $i$ are updated by
	\begin{equation}\label{x-legal}
		x_i'=\left\{
		\begin{array}{rcl}
			\mathop{max}\limits_{\forall v_{j} \in I_i} (x_{j}'+w_{j}),	&	&	\mbox{if } I_i \neq \emptyset \\
			0,	&	&	\mbox{otherwise;}
		\end{array} \right.
	\end{equation}
	\begin{equation}\label{y-legal}
		y_i'=\left\{
		\begin{array}{rcl}
			\mathop{max}\limits_{\forall v_{j} \in J_i} (y_{j}'+h_{j}),	&	&	\mbox{if } J_i \neq \emptyset \\
			0,	&	&	\mbox{otherwise.}
		\end{array} \right.
	\end{equation}

	\begin{algorithm}[htb]
		\caption{$(\bm{x}^*,\bm{y}^*)=Graph(\bm{x},\bm{y})$}\label{alg:Graph}
		\KwIn{$(\bm{x},\bm{y})$;}
		\KwOut{$(\bm{x}^*,\bm{y}^*)$;}
		Sorting all modules according to the $x$-coordinates of the bottom-left corner and denote them as $\{v_1,v_2,\dots,v_n\}$\;
		\For{$i \leftarrow 1$ to $n$}
		{update $x_i'$ and $\bm{x}^*$ according to formula (\ref{x-legal});
		}
		Sorting all modules according to the $y$-coordinates of the bottom-left corner and denote them as $\{v_1,v_2,\dots,v_n\}$\;
		\For{$i \leftarrow 1$ to $n$}
		{update $y_i'$ and $\bm{y}^*$ according to formula (\ref{y-legal});
		}
	\end{algorithm}

\section{The Floorplanning Algorithm Based on the Golden Section Strategy}\label{sec:FA-GSS}
While the analytical optimization method is applied to the floorplanning problem without fixed-outline, it is a challenging task to minimize the floorplan area.	In this paper, we proposed a floorplanning algorithm based on the golden section strategy (FA-GSS), where minimization of the floorplan area is achieved by consecutively narrowing the contour of fixed outline.
	
	\begin{algorithm}[htb]
		\caption{FA-GSS}\label{alg:GSS}
		\KwIn{$f(\bm{x},\bm{y},\bm{r})$, $\widetilde{f}(\bm{x},\bm{y}),\bm{r}$.}
		\KwOut{Optimal solution $(\bm{x}^*,\bm{y}^*)$ and corresponding rotation strategy $\bm{r}^*$.}
		initialize $\bm{Q}(0)$ according to formula (7), and sample to generate $\bm{P}(0)$\;
		initialize $\bm{X}$ and $\bm{Y}$, step control parameter $s$\;
		set the best solution in $\bm{P}(0)$ is $\bm{p}^*$, the corresponding distribution matrix is $\bm{q}^*$, and the module coordinates are $(\bm{x}^*,\bm{y}^*)$\;
		$\lambda_0 \leftarrow 1$, $\mu_0 \leftarrow 10$, $k_{\mbox{max}} \leftarrow 50$, $t\leftarrow 1$\;
		initialize the maximum whitespace ratio $\gamma_{\mbox{max}}$ and  minimum whitespace ratio $\gamma_{\mbox{min}}$\;
		\While{$\gamma_{\mbox{max}}-\gamma_{\mbox{min}}<\epsilon$}
		{
			$\alpha \leftarrow 1$, $\lambda \leftarrow 20$, $\mu \leftarrow 100$, $\gamma_{\mbox{m}}=0.618*(\gamma_{\mbox{max}}-\gamma_{\mbox{min}})+\gamma_{\mbox{min}}$\;
			calculate the width $\bm{W}^*$ and height $\bm{H}^*$ of the fixed profile\;
			\While{termination-condition 2 is not satisfied}
			{$\bm{Q}'(t)=OrthExpQ(\bm{Q}(t-1),\bm{P}(t-1))$\;
				$\bm{P}'(t)=SampleP(\bm{Q}(t),\bm{P}(t-1))$\;
				$(\bm{P}(t),\bm{X},\bm{Y},s)=UpdateXY(\bm{P}'(t),\bm{X},\bm{Y},s)$\;
				$\bm{Q}(t)=RefineQ(\bm{P}'(t),\bm{P}(t),\bm{Q}'(t))$\;
				Let's take the best solution for the current $\bm{X}$, $\bm{Y}$, and call it $\bm{x}'$, $\bm{y}'$\;
				$t=t+1$, update $\bm{p}^*$, $\bm{q}^*$, $(\bm{x}^*,\bm{y}^*)$, $k_{\mbox{max}}=35$;}
			\eIf{$\widetilde{f}(\bm{x}',\bm{y}')=0$}{$\gamma_{\mbox{max}}$=$\gamma_{\mbox{m}}$;}{$\gamma_{\mbox{min}}$=$\gamma_{\mbox{m}}$;}
		}
	\end{algorithm}

Minimization of the floorplan area $S(\bm x,\bm y, \bm r)$ is equivalent to minimizing the blank ratio
	\begin{equation}\label{obj:gamma}
		\gamma=\frac{S(\bm x,\bm y, \bm r)-A}{A}*100\%,
	\end{equation}
where $A$ is the sum of areas of all modules. As presented in Algorithm \ref{alg:GSS}, we use the golden section strategy to continuously reduce the area of the fixed contour. Given the initial white rate $\gamma_{{max}}$ and $\gamma_{{min}}$, where the fixed-outline floorplanning is feasible for $\gamma_{{max}}$ but infeasible for $\gamma_{{min}}$, we set $$\gamma_{{m}}=0.618*(\gamma_{{max}}-\gamma_{{min}})+\gamma_{{min}}.$$ If a legal layout can be obtained for $\gamma_{{m}}$, then $\gamma_{{max}}$=$\gamma_{{m}}$; otherwise, we set $\gamma_{{min}}$=$\gamma_{{m}}$. Repeat the section process  until $\gamma_{{max}}-\gamma_{{min}}<\epsilon$.
	
\section{Experimental Results and Analysis}\label{sec:Num}
To verify the performance of the proposed algorithm, we conducted experiments on the well-known test benchmark GSRC. For all test circuits, the I/O pads are fixed at the given coordinates, and the modules of all circuits are hard modules. All experiments are developed in C++ programming language program, and run in Microsoft Windows 10 on a laptop equipped with the AMD Ryzen 7 5800H @ 3.2GHz and 16GB system memory.
\subsection{Wirelength Optimization with Fixed-outline Constraints
}
We first test the performance of FFA-CD on the fixed-outline cases. It is compared with the well-known open source layout planner Parquet-4.5~\cite{clar:eke17}, where the floorplan is represented by the B*-tree and the simulated annealing algorithm to solve the combinatorial optimization model of floorplanning.

According to the given aspect ratio $R$, the width $W^*$ and height $H^*$ of the fixed contour are calculated as~\cite{clar:eke16}
\begin{equation}
	W^*=\sqrt{(1+\gamma)A/R},H^*=\sqrt{(1+\gamma)AR},
\end{equation}
where $A$ is the summed area of all modules, and $\gamma$ is the white rate defined in (\ref{obj:gamma}).
The experiment set the white rate as $\gamma=15\%$, the aspect ratio R as 1, 1.5, 2, and the population number as 5. For different aspect ratios, each experiment was independently run 10 times, and the results were shown in Table \ref{Tab1}. 

Numerical results demonstrate that FFA-CD outperforms Parquet-4.5 on cases with more than 50 modules, but runs a bit slow for some of the small cases, which is attributed to the compact floorplan of Parquet-4.5. The combinatorial floorplan implemented by Parquet-4.5 could lead to smaller HPWL and shorter runtime, but its performance would degrade significantly while the problem size increases.
The iteration mechanism based on CSA ensures that FFA-CD can explore the floorplan space more efficiently. At the same time, DEA-PPM is introduced to explore the rotation strategy, which increases the flexibility of the floorplan and greatly improves the success rate of small-scale problems.
Consequently, the success rate of FF-CD was better than or equal to Parquet-4.5 for all cases. Meanwhile, better results on wirelength and tuntime is obtained  in several different aspect ratios for the larger-scale cases (n50-n100).

\begin{table}[htb]\centering
\caption{Performance comparison for the fixed-outline cases of GSRC test problems.}\label{Tab1}
\begin{tabular}{*{8}{c}}
	\toprule
	\multirow{2}*{GSRC} & \multirow{2}*{R} & \multicolumn{3}{c}{Parquet-4.5} & \multicolumn{3}{c}{FFA-CD} \\
	\cmidrule(lr){3-5}\cmidrule(lr){6-8}
	& & SR(\%) & HPWL & CPU(s) & SR(\%) & HPWL & CPU(s) \\
	\midrule
	\multirow{3}*{n10} & 1.0 & 60 & \textbf{55603} & \textbf{0.04} & 100 & 55774 & 0.11 \\
	& 1.5 &60 & \textbf{55824} & \textbf{0.04} & 100 & 56696 & 0.20\\
	& 2.0 &80 & 58247 & \textbf{0.04} & 90 & \textbf{58236} & 0.31\\
	\midrule
	\multirow{3}*{n30} & 1.0 & 100 & 172173 & \textbf{0.28} & 100 & \textbf{160208} & 0.41 \\
	&1.5 & 90 & 173657 & 0.34 & 100 & \textbf{164237} & \textbf{0.28}\\
	&2.0 & 100 & 174568 & \textbf{0.32} & 100 & \textbf{166133} & 0.54\\
	\midrule
	\multirow{3}*{n50} & 1.0 & 100 & 209343 & 0.68 & 100 & \textbf{185793} & \textbf{0.55} \\
	&1.5 & 100 & 211591 & 0.79 & 100 & \textbf{189878} & \textbf{0.41}\\
	&2.0 & 100 & 208311 & 0.78 & 100 & \textbf{195398} & \textbf{0.71}\\
	\midrule
	\multirow{3}*{n100} & 1.0 & 100 & 334719 & 2.10 & 100 & \textbf{293578} & \textbf{0.89} \\
	&1.5 & 100 & 340561 & 2.26 & 100 & \textbf{300079} & \textbf{1.05}\\
	&2.0 & 100 & 347708 & 2.26 & 100 & \textbf{308811} & \textbf{1.02}\\
	\midrule
	\multirow{3}*{n200} & 1.0 & 100 & 620097 & 9.03 & 100 & \textbf{521140} & \textbf{2.38} \\
	&1.5 & 100 & 625069 & 9.07 & 100 & \textbf{529918} & \textbf{2.53}\\
	&2.0 & 100 & 649728 & 9.24 & 100 & \textbf{541565} & \textbf{2.71}\\
	\midrule
	\multirow{3}*{n300} & 1.0 & 100 & 768747 & 19.08 & 100 & \textbf{588118} & \textbf{3.73} \\
	&1.5 & 100 & 787527 & 19.16 & 100 & \textbf{606548} & \textbf{3.85}\\
	&2.0 & 100 & 847588 & 19.63 & 100 & \textbf{626658} & \textbf{4.21}\\
	\bottomrule
\end{tabular}
\end{table}

	
\subsection{Minimization of Wirelength and Area without Fixed-outline Constraints}
For layout planning problems without fixed contour constraints, FA-GSS is used to optimize the wirelength and area. The proposed FA-GSS is compared with Parquet-4.5 and the Hybrid Simulated Annealing Algorithm (HSA)~\cite{clar:eke18}, where the population size is set as 5, and we get $\epsilon=0.2\%$.
Due to the different magnitude of wirelength and area, the cost function to minimized for the floorplanning problem without fixed outline is taken as
\begin{equation}
	Cost=0.5*\frac{W}{W_{{min}}}+0.5*\frac{S}{S_{{min}}},
\end{equation}
where $W_{{min}}$ and $S_{{min}}$ are the minimum values of $W$ and $A$, respectively.

The results in Table \ref{Tab1} show that all examples obtain better wirelength and shorter time when the aspect ratio is 1. So, we take $R=1$ in FA-GSS for all test cases. For benchmarks in GSRC, the average $Cost$, and runtime (CPU) of  ten independent runs are collected in Table 2.

\begin{table}\centering
	\caption{Performance comparison for the GSRC test problems without fixed-outline constraints.}\label{Tab2}
	\begin{tabular}{*{7}{c}}
		\toprule
		\multirow{2}*{GSRC} & \multicolumn{2}{c}{Parquet-4.5} & \multicolumn{2}{c}{HAS} & \multicolumn{2}{c}{FA-GSS} \\
		\cmidrule(lr){2-3}\cmidrule(lr){4-5}\cmidrule(lr){6-7}
		& $Cost$ & CPU(s) & $Cost$ & CPU(s) & $Cost$ & CPU(s)\\
		\midrule
		\multirow{1}*{n10} & 1.0885 & \textbf{0.03} & 1.0799 & 0.11 & \textbf{1.0688} & 0.17 \\
		\midrule
		\multirow{1}*{n30} & 1.1040 & \textbf{0.19} & \textbf{1.0881} & 0.86 & 1.0959 & 0.69 \\
		\midrule
		\multirow{1}*{n50} & 1.0871 & \textbf{0.47} & 1.0797 & 2.15 & \textbf{1.0750} & 1.29 \\
		\midrule
		\multirow{1}*{n100} & 1.1034 & \textbf{1.61} & 1.1040 & 7.94 & \textbf{1.0648} & 3.53 \\
		\midrule
		\multirow{1}*{n200} & 1.1301 & \textbf{6.23} & 1.1628 & 37.7 & \textbf{1.0713} & 8.96 \\
		\midrule
		\multirow{1}*{n300} & 1.1765  & \textbf{12.87} & 1.2054 & 78.21 & \textbf{1.0715} & 15.13 \\
		\bottomrule
	\end{tabular}
\end{table}

The experimental results show that FA-GSS outperforms both Parquet-4.5 and HAS except for the n30 case. Although FA-GSS runs a bit slower than Parquet-4.5 when they are tested by the n30 case, FA-GSS has the smallest rate of increase in run time as the module size increases. This means that FA-GSS is expected to achieve excellent results on larger circuits.

\section{Conclusion}\label{sec:Con}
In this paper, we formulate the flooplanning problem of VLSI as a mixed-variable optimization problem, where the discrete variables represent module orientations and the coordinates of modules are incorporated by continuous variables. Then, the DEA-PPM is introduced to get the module orientation, and coordinate variables are optimized by the CSA.  Experimental results show that the proposed FFA-CD and FA-GSS, respectively developed for floorplanning problems with and without fixed-outline, can generally outperforms the floorplanning algorithms designed based on the B*-tree and the simulated annealing. Attributed to their low time complexity, the proposed algorithms are expected to address large-scale floorplanning problems effectively.

\section*{Acknowledgement}
This research was partially supported by the National Key R\& D Program of China (No.2021ZD0114600) and the Guiding Project of Scientific Research Plan of Hubei Provincial Department of Education (No. B2022394).

\end{document}